\documentclass[runningheads]{llncs}
\usepackage{graphicx}
\begin{document}

\title{Learning Interpretable Features via Adversarially Robust Optimization}
%
%
\author{Ashkan Khakzar \inst{1} \and
Shadi Albarqouni\inst{1} \and
Nassir Navab\inst{1,2}}
%
\authorrunning{A. Khakzar et al.}
%
\institute{Technical University of Munich, Munich Germany \and
Johns Hopkins University, Baltimore MD USA}
\maketitle              
\begin{abstract}

Neural networks are proven to be remarkably successful for classification and diagnosis in medical applications. However, the ambiguity in the decision-making process and the interpretability of the learned features is a matter of concern. In this work, we propose a method for improving the feature interpretability of neural network classifiers. 
Initially, we propose a baseline convolutional neural network with state of the art performance in terms of accuracy and weakly supervised localization.
Subsequently, the loss is modified to integrate robustness to adversarial examples into the training process. 
In this work, feature interpretability is quantified via evaluating the weakly supervised localization using the ground truth bounding boxes. Interpretability is also visually assessed using class activation maps and saliency maps.
The method is applied to NIH ChestX-ray14, the largest publicly available chest x-rays dataset. We demonstrate that the adversarially robust optimization paradigm improves feature interpretability both quantitatively and visually. 
\keywords{Interpretability  \and Medical Imaging \and Adversarial Training.}
\end{abstract}
%
%
%
%
%
\section{Introduction}
Deep learning methods have shown great promise in medical diagnosis \cite{greenspan2016guest}. Specifically, after the release of NIH ChestX-ray dataset, deep learning based methods achieved high performances on chest x-ray classification hitherto unprecedented on large scale \cite{wang2017chestx,rajpurkar2017chexnet,li2018thoracic,yao2017learning}. Despite these promising accomplishments, the previously proposed methods for chest x-ray classification do not show adequate feature interpretability, while similar methods achieve higher feature interpretability on computer vision datasets \cite{zhou2016learning}. Feature interpretability
is critical in the clinical setting, as it helps explain why a diagnostic decision is made by the classifier \cite{biffi2018learning,miotto2017deep}. Therefore it is also critical that the methods proposed for medical image diagnosis achieve high scores in interpretability metrics.

%
Several works consider interpretability of neural network classifiers.
Rajpurkar et al. \cite{rajpurkar2017chexnet} propose a classification model for Pneumonia detection on NIH ChestX-ray14 dataset and visualize Class Activation Maps (CAMs) \cite{zhou2016learning} to show interpretability of features used for Pneumonia prediction. They also propose a multi-label classification model with high classification accuracy for all classes in the dataset, however do not evaluate its interpretability.
Wang et al. \cite{wang2017chestx} propose a unified weakly supervised multi-label image classification and disease localization framework. The localization, which is based on CAMs, serves as a metric for evaluating the interpretability of the features. Their proposed weighted cross entropy scheme enforces the model to learn interpretable features in the highly imbalanced (in terms of positive and negative examples) NIH ChestX-ray dataset. However, the localization results do not keep pace with the classification results.
Li et al. \cite{li2018thoracic} improve the localization on the same dataset significantly by incorporating the annotated bounding boxes into training. Although this is an intuitive method to enforce the model to learn explainable features, lack of annotated data hinders the use of this approach.
Biffi et al. \cite{biffi2018learning} propose a method based on convolutional generative neural networks for designing models with interpretable features and the method is applied to the classification of cardiovascular diseases. The method enforces interpretability by design and limits the architecture of the neural network to only generative ones. Therefore, an approach that would enforce interpretability on all high performing models (and not just a specific architecture) is appreciated.

%
%
It is postulated that feature representations learned using robust training capture salient data characteristics~\cite{tsipras2018robustness}. Adversarially robust optimization is introduced as a method for robustness against adversarial examples in \cite{goodfellow2014explaining,madry2017towards}.
In this work, we improve the interpretability of the state of the art neural network classifiers via adversarially robust optimization. The work tries to steer the models toward learning features that are more semantically relevant to the pathologies in the classification problem. Initially, we propose a baseline neural network classifier based on the state of the art. Then we modify its loss to adversarially robust loss and measure the improvement in terms of interpretability and classification accuracy.
To evaluate the feature interpretability of the proposed solution, its localization accuracy is measured. Moreover, CAMs and saliency maps \cite{simonyan2013deep} are presented for visual evaluation.
\begin{figure}[!t]
\includegraphics[width=\textwidth,keepaspectratio]{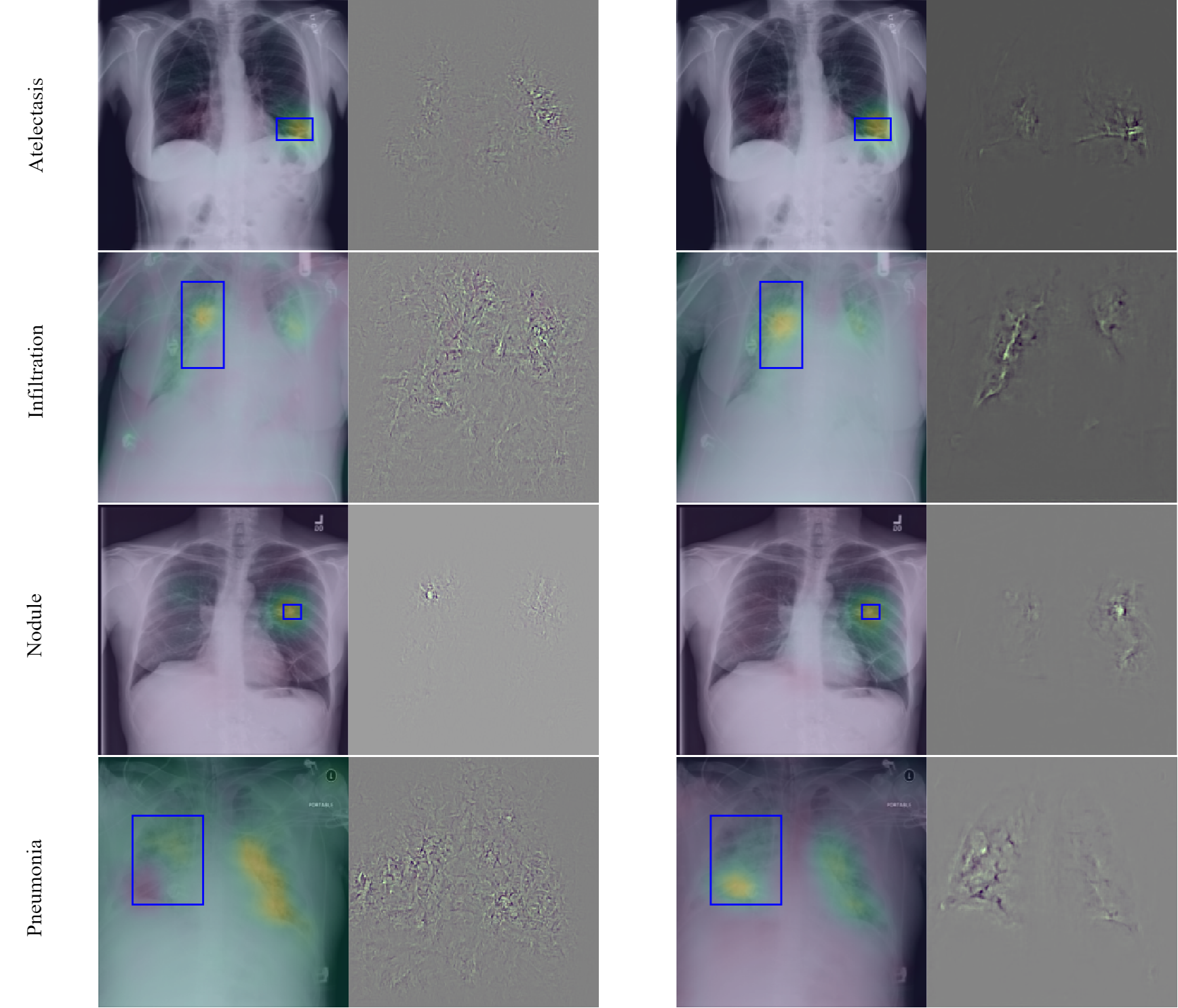}
\caption{Visualization of CAMs (overlayed on the input image) and saliency maps. (Left) our proposed baseline. (Right) our adversarially robust optimization method ($\epsilon = 0.005$). Blue boxes are ground truth annotation. (Images are best seen in electronic form) }
\label{fig:visualizations}
\end{figure}
\section{Methodology}
\subsection{Baseline Model}
Given a database $X =\{x^{(1)}, ..., x^{(N)}\}$ of $N$ X-ray images, and their corresponding labels $Y=\{y^{(1)}, ..., y^{(N))}\}$, where $y = [y_{1},...,y_{j},...,y_{C}]$ and $y_{j}\in\{0,1\}$ with $C$ being the number of classes, our aim is to train a model $\hat{y} = f_\theta(x)$, where $\hat{y}$ is the predicated label and $\theta$ denotes the parameters of the model. The loss to be minimized is binary cross entropy loss, and for each input example $x$ is defined as:
\begin{equation}
\mathcal{L}(f_\theta(x), y) = - \sum_{j} \beta y_{j}\, \log (\hat{y_{j}}) + (1-y_{j})\, \log (1-\hat{y_{j}}) 
\label{eq:loss}
\end{equation}
where $\beta$ is a weighting factor to balance the positive labels, and defined as the ratio of the number of negative labels to the number of positive labels in a batch. 
\subsection{Adversarially Robust Optimization}
This work aims at improving the learned feature representations of neural network classifiers through training models that are robust against adversarial examples.
We view adversarial examples and robustness from the perspective of optimization. 
Given the loss formulated in equation \ref{eq:loss}, adversarial examples are perturbed inputs that try to maximize the loss
\begin{equation}
\max _{\delta \in \Delta } \mathcal{L}(f_{\theta }(x+\delta ),y)
\label{eq:adversarial_example}
\end{equation}
where $\delta$ is the perturbation and $\Delta$ defines the allowed perturbation \cite{goodfellow2014explaining}.
In order to make models robust against these adversarial examples, the loss is modified to a min-max problem so that it incorporates robustness as an objective \cite{madry2017towards}: 
\begin{equation}
\min_{\theta}\max _{\delta \in \Delta } \mathcal{L}(f_{\theta }(x+\delta ),y)
\label{eq:minmax}
\end{equation}
The approach for solving the optimization problem is to repeatedly find input perturbations $\delta$ by solving the inner maximization, and then update the model parameters $\theta$ to reduce the loss on these perturbed inputs. 
It is not necessary for the inner maximization to be solved exactly, and an approximate lower bound could lead to a reasonable solution for the min-max problem \cite{madry2017towards}.
Our purpose for robust optimization is not only having a high performance against adversarial attacks, but also steering the model towards learning more interpretable features. We choose a $\Delta$ and optimization method for Eq. \ref{eq:adversarial_example} that is computationally reasonable, and we show that the model still learns robust features. 
%
%
%
%
\section{Experiments}
\subsection{Dataset}
We evaluate our method on the NIH chestX-ray14 dataset \cite{wang2017chestx}, which is the largest publicly available chest x-ray dataset to date and includes 112,120 frontal-view X-ray images of 30,805 unique patients in $1024\times1024$ resolution. Each image has fourteen labels associated with it, each corresponding to common thoracic pathologies.
We use the train/test split provided with the dataset in its latest update, i.e. from the entire dataset, 25596 images are in the test set. The rest are split to training (90\%) and validation (10\%) sets. In the test set, 880 images have bounding box annotations of at least one pathology. Annotations exist only for 8 out of 14 pathologies.

\subsection{Baseline Model}
In previous works \cite{wang2017chestx,rajpurkar2017chexnet,li2018thoracic}, the feature maps of the last layer are of low resolution and they are used for generating the CAMs. These Low-resolution CAMs are not able to localize pathologies such as Nodule that are small in size.
Hence, it is intuitive to modify the CNN to have larger feature maps in the last layer.
Therefore we adopt a densely connected convolutional neural network \cite{huang2017densely}, DenseNet-121, and remove the dense-blocks 3 and 4 and their corresponding transition layers (2 and 3) in order to get higher resolution feature maps in the last layer.
The aforementioned neural network serves as our baseline model and its classification accuracy and interpretability evaluation are depicted in Fig. \ref{Fig:auc_all} and Fig. \ref{fig:localization_iou_comparison} respectively.

In all our experiments, the networks are trained using stochastic gradient descent with a learning rate of 0.01 and a momentum of 0.9. We use a batch size of 32 and do not use weight decay and dropout. The training is continued until the validation loss (Eq. \ref{eq:loss}) diverges, and the model with the smallest validation loss is used.
\subsection{Interpretability Evaluation}\label{interpret_eval}
Weakly-supervised localization accuracy is measured for each classification model and is used as a proxy for evaluating interpretability of the classification model. Localization is evaluated using intersection over union (IoU) between the thresholded CAM and the bounding box. The localization is correct when the IoU is greater than a certain threshold T(IoU). Localization accuracy is calculated for several values of T(IoU). 

We do not generate bounding boxes from the thresholded CAMs and follow the same approach proposed by Li et al. \cite{li2018thoracic}, where the feature maps are directly evaluated using IoU metric without any bounding box generation. First, we upsample the CAMs via bilinear interpolation to the input image size that is $224\times224$ and then scale their values to a range (we used [0 255]) and subsequently threshold them by value. The method does not depend on further post-processing and bounding box generation approaches, thus directly evaluates the feature maps. We choose the thresholding value differently for each class based on its resulting localization performance on a validation set. The validation set is selected from the annotated images (from 880 images) in the test set. 20\% (of 880 images) is chosen for finding the thresholding value and we perform 5-fold cross validation for correct evaluation.
%

%
\subsection{Adversarially Robust Optimization}
The min-max optimization in Eq. \ref{eq:minmax} is solved iteratively by solving the inner maximization and then the outer minimization, hence a method that requires several update steps for solving the inner maximization makes the solution computationally expensive. Many methods have been proposed in the literature for finding approximate solutions to the inner maximization problem. We use the FGSM \cite{goodfellow2014explaining} method as it requires only one update step for finding a local maxima.
If we start solving the min-max problem (Eq. \ref{eq:minmax}) from a network not yet trained on the dataset, the adversarial examples make it hard for the network to learn the features of the dataset. Therefore, we initially train the network without adversarial loss, and after convergence, we continue training with the adversarial loss in Eq. \ref{eq:minmax}. We observed that it is also helpful for training convergence, not to perturb all examples during training. Hence we only perturb half \cite{tsipras2018robustness} of the input examples in each epoch during training with the adversarial loss (Eq. \ref{eq:minmax}).
The amount of perturbation $\delta$ allowed for the FGSM method is defined by $\epsilon$. FGSM finds a local maxima in Eq. \ref{eq:adversarial_example} limited by the allowed perturbation set  $\Delta = l_{\infty}$. Several values for $\epsilon$ are chosen in the experiments in order to see the effect of the amount of perturbation during training on the learned features of the network Fig. \ref{fig:progression}). As can be seen in Fig. \ref{fig:progression}, $\epsilon = 0.005$ results in more interpretable saliency maps, hence for quantitative analysis and comparison of the robust model with the baseline we use $\epsilon = 0.005$.
\begin{figure}[t]
\includegraphics[width=\textwidth,keepaspectratio]{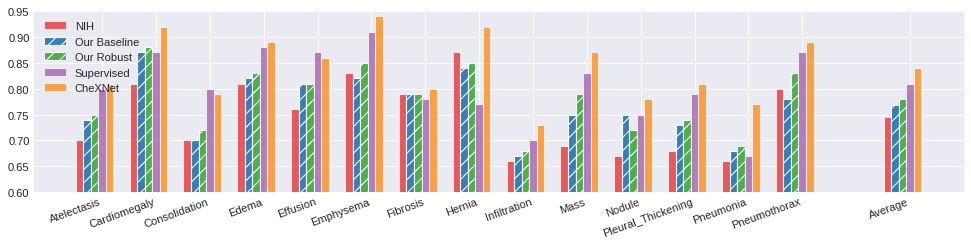}
\caption{Classification accuracies (AUC of ROC curve) for our proposed models and state of the art.}
\label{Fig:auc_all}
\end{figure}
\begin{figure}[!t]
\includegraphics[width=\textwidth,keepaspectratio]{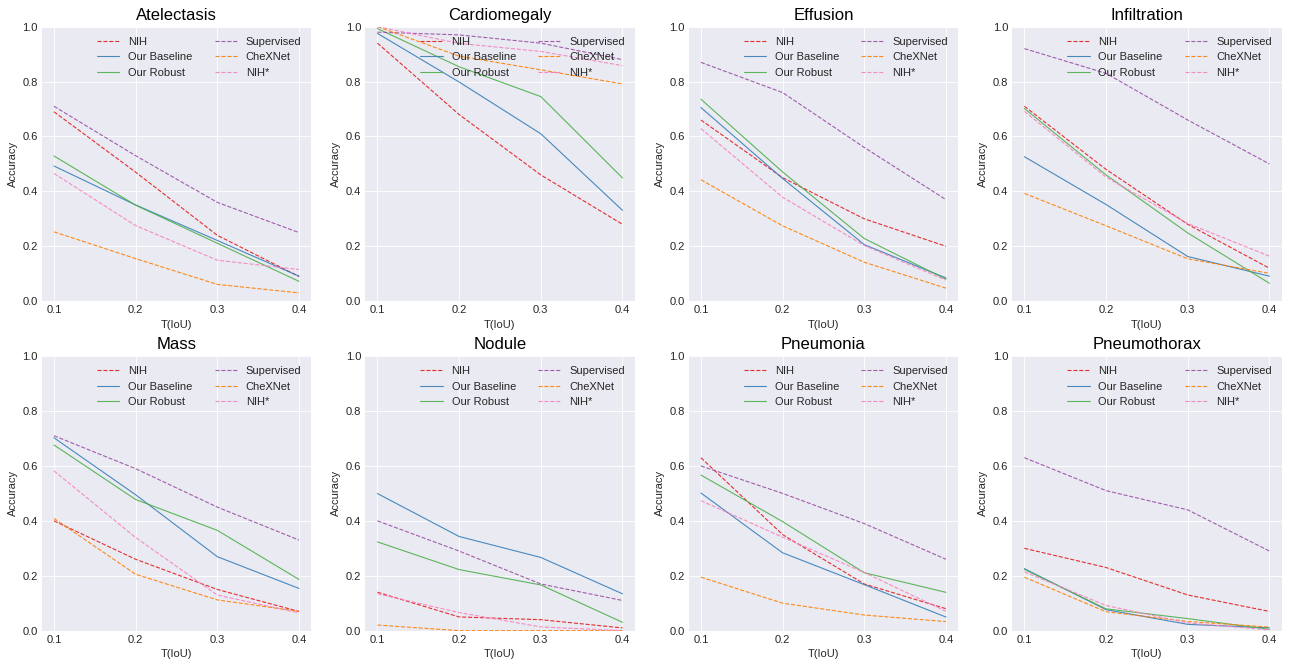}
\caption{Localization accuracy of state of the art models (dashed lines) \cite{wang2017chestx,rajpurkar2017chexnet,li2018thoracic} and our proposed models (Baseline and Robust). The horizontal axis represents the T(IoU) used for computing the localization accuracy}
\label{fig:localization_iou_comparison}
\end{figure}
\section{Results and Discussion}
In this section and the Figures \ref{Fig:auc_all} and \ref{fig:localization_iou_comparison} we refer to the work of Wang et al. \cite{wang2017chestx} as NIH method, Rajpurkar et al. \cite{rajpurkar2017chexnet} as CheXNet, Li. et al. \cite{li2018thoracic} as Supervised and our method that is based on adversarially robust optimization as Robust method.
\subsection{Baseline Model vs. State Of the Art}\label{eval_baseline}
CheXNet method achieves the highest AUC (Fig. \ref{Fig:auc_all}). However, it has the lowest localization accuracy (Fig \ref{fig:localization_iou_comparison}), indicating that the model's learned features do not align well with the pathologies. The high AUC and lack of interpretable features of CheXNet can be attributed to its unweighted binary cross entropy loss where the imbalance between positive and negative examples is ignored.

The Supervised method uses 80\% of annotated images during training, hence it achieves the highest localization accuracy. We report it as an upper bound for the localization accuracy, and it cannot be fairly compared with other methods since they are only trained on labels. However, our baseline surpasses the Supervised method in Nodule localization as it generates higher resolution CAMs.

For localization accuracy, NIH uses a different evaluation method than ours (section \ref{interpret_eval}). NIH method uses an ad-hoc CAM thresholding and bounding box generation approach. Therefore, in order to compare our baseline fairly with NIH, we implemented NIH (using ResNet50 without transition layer) method and evaluated it using the procedure in section \ref{interpret_eval}, which is shown as NIH* in Fig. \ref{fig:localization_iou_comparison}. 
\subsection{Robust Model vs. Baseline}\label{eval_robust}
\textbf{Quantitative Evaluation:} 
Our Robust model (trained with $\epsilon = 0.005$) shows improvement in localization accuracy for Cardiomegaly, Pneumonia, and Infiltration, yields lower accuracy for the Nodule class and comparable (still higher) results for the rest.
Nevertheless, while quantitative results provide a means for measuring feature interpretability of the model on the entire test set, visually explainable features may still be essential for the clinical community.
\begin{figure}[t]
\includegraphics[width=\textwidth,keepaspectratio]{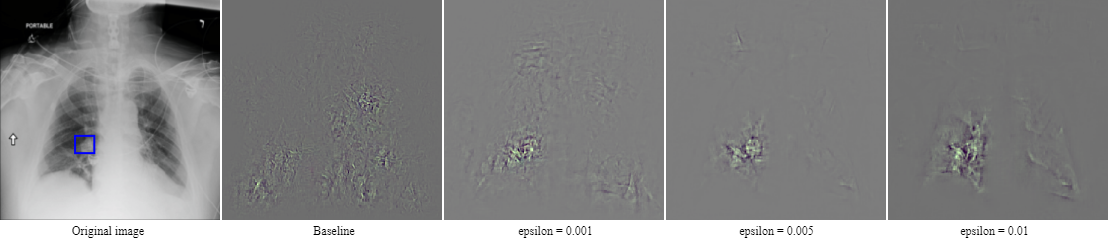}
\caption{Effect of increasing the perturbation $\delta$ ($\delta = |\epsilon|$) used in our robust method on saliency maps. The blue box denotes the ground truth bounding box for Mass. (Images are best seen in electronic form)}
\label{fig:progression}
\end{figure}
\\
\textbf{Visual Evaluation:}
The robust model yields significantly more interpretable gradients with respect to the input image as seen in saliency maps in Fig. \ref{fig:visualizations}. These are vanilla saliency maps \cite{simonyan2013deep}, the gradients are only clipped by three standard deviations and scaled to [0 1] and no further processing (e.g. smoothing) is performed. 
The effect of increasing the amount of perturbation $\epsilon$ during adversarial robust optimization on visual interpretability is presented in Fig. \ref{fig:progression}. It can be seen that increasing the amount of perturbation during training steers the model toward focusing on the most salient feature of the image.
It is also interesting in future research to study the effects of other perturbation sets $\Delta$ such as rotations on feature interpretability.
\section{Conclusion}
In this work, we demonstrated that adversarially robust optimization improves the feature interpretability of neural network classifiers both quantitatively and visually. Saliency maps of our adversarially trained models show significantly more interpretable features. The method does not have any dependency on the neural network architecture and the dataset. 
We also demonstrated that evaluating the model only using classification accuracy is not reliable since the high accuracy of a model could be due to its reliance on features that are not relevant to the pathologies.

\subsubsection{Acknowledgement}
We would like to thank Siemens Healthineers  for their financial support.


\end{document}